# A Probabilistic Transmission Expansion Planning Methodology based on Roulette Wheel Selection and Social Welfare

[1]Neeraj Gupta, [2]Rajiv Shekhar, and [1]Prem Kumar Kalra

*Abstract:* **A new probabilistic methodology for transmission expansion planning (TEP) that does not require a priori specification of new/additional transmission capacities and uses the concept of social welfare has been proposed. Two new concepts have been introduced in this paper: (i) roulette wheel methodology has been used to calculate the capacity of new transmission lines and (ii) load flow analysis has been used to calculate expected demand not served (EDNS). The overall methodology has been implemented on a modified IEEE 5-bus test system. Simulations show an important result: addition of only new transmission lines is not sufficient to minimize EDNS.**

## 1. INTRODUCTION

Nowadays, the need for appropriate planned power systems to reduce generation cost, minimize the consumer cost and improve the quality of the power supply has become imperative [1]-[3]. As a result, transmission expansion planning (TEP) is gaining more significance. A sub-optimally planned transmission network may lead to unutilized generation capacities, demand not served, and even over investment. This is particularly important, for example, in India where the transmission capacity has to be increased significantly, primarily to keep pace with the expected 11% growth in generation [4]. Previous studies [5]-[8] have identified important issues pertaining to network planning: (i) optimal locations of new transmission lines, (ii) up-gradation of existing transmission lines, and (iii) optimal capacity of the proposed transmission lines.

[1] N.Gupta and P. K. Kalra are with the Department of Electrical Engineering, Indian Institute of Technology Kanpur 208016, UP, India (e-mail: ngtaj@iitk.ac.in).

[2] R. Shekhar is with the Department of Material Science and Engineering, Indian Institute of Technology Kanpur, 208016, UP, India (e-mail: vidtan@iitk.ac.in).

Traditionally, the deterministic N-1/N-2 contingency planning methodology has been used for TEP. However, it cannot account for the probabilistic nature of generation and transmission equipment failures, resulting in an under/ over designed transmission system. A powerful methodology used in "stochastic" TEP has been a combination of i) optimization, ii) probability theory, (iii) Monte Carlo simulation (*MCS*), and iv) graph theory [3], [9]-[11]. Here, new transmission lines were identified based on the optimization of the total costs.

Su and Li [12], [13] minimized the sum of *EDNS* and transmission investment costs to determine new transmission lines. *MCS* was used to characterize the stochastic nature of the transmission lines and generation failure, while graph theory, or more specifically, the min-cut max-flow algorithm [14], was used to calculate *EDNS*. The approach of Choi et al. [15] was similar to that of Su and Li [12] except for the fact that reliability criteria were used as a constraint, while minimizing the investment costs. Akbari et al. [16] presented a TEP methodology using probabilistic selection of loads. However, they did not account for the multiple stochastic failures of generation and transmission equipments.

Recent papers on TEP [17], [18] have incorporated load-flow analysis -- instead of min-cut max-flow approach – to compute *EDNS/ EENS* using the load-curtailment strategy. This process requires multiple load flow calculations to compute *DNS* for one network configuration.

Consequently, deterministic contingency criteria were used instead of probabilistic *MCS* contingency criteria. In their general review of TEP methodologies, Shahidepour [3], Latorre et al. [9] and Lee et al. [10] have advocated the use of i) social welfare, ii) scenario analysis, and iii) trade-off between economics and reliability of the power systems for designing networks.

The approach followed by previous investigations [1]-[18] have several shortcomings. First, the capacity of all possible new transmission lines were specified *a priori*. Second, the network modeling and related computations in graph theory do not follow electrical laws. In fact, application of DC-load flow and suggested *EDNS* calculation approach on the optimized network presented in [12], showed that the *EDNS* was 165 MW compared to 48 MW calculated using the min-cut max-flow method. Third, the scenario analysis based on the load curve should be integrated with TEP to avoid overinvestment[3]. Fourth, the economic dispatch of generators is not followed in the graph theory approach. Fifth, the concept of social welfare has not been incorporated. For example, non-zero *EDNS* also implies unutilized generation capacity or expected generation not served (*EGNS*), the cost of which must be accounted for in TEP. Similarly, the cost associated with wheeling losses (*WL*) should be incorporated.

The major objective of this investigation has been to develop a probabilistic methodology for TEP which captures the realities of a transmission system, as mentioned in the previous paragraph. The proposed methodology is illustrated through a case study. The paper is organized in the following manner. Section *2* presents the TEP methodology in detail, which includes discussions on the load flow analysis for calculation of *EDNS* and roulette wheel methodology for determining the capacities of new transmission lines. The detailed algorithm of the proposed methodology is given in section *3*. Results of the implementation of the TEP methodology in a modified IEEE 5-bus system are given in Section *4*.

---

[3] For deterministic planning such as N-1, it may be reasonable to use the peak load level to justify transmission requirement due to the less stringent consideration of system contingency scenarios. However, for probabilistic planning through *MCS*, not accounting for the LDC may result in extreme values of EDNS and EGNS.

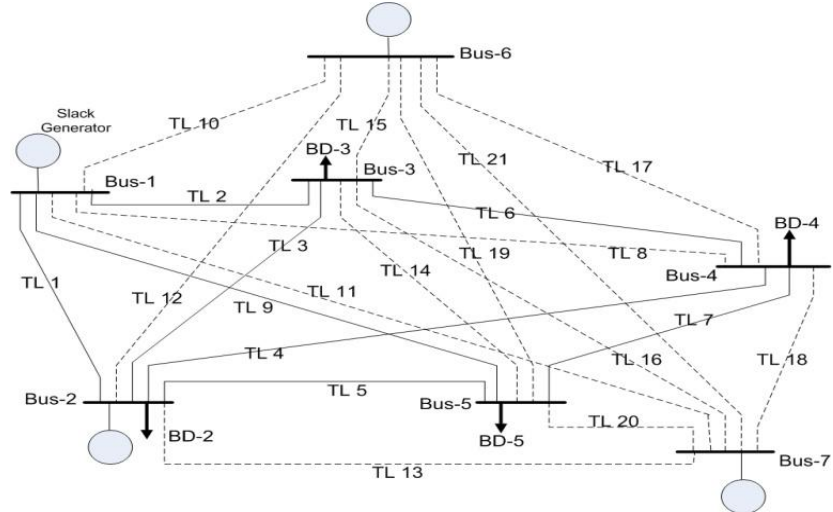

Fig. 1. Schematic diagram of the IEEE 5-bus power system, the dotted lines represent the probable new transmission lines.

## 2. METHODOLOGY

The proposed methodology involves: (i) minimization of the sum of the investment, *EDNS, EGNS* and *EWL* costs to put together economic and reliability analysis on a single platform, (ii) incorporating contingency analysis using *MCS*, (iii) merit order dispatch of generators, (iv) DC-load flow analysis, (v) Kirchhoff's law to determine *EDNS/EGNS* in the electrical network, and (vi) roulette wheel selection method to calculate the optimum capacity of the transmission lines. Implicit in the proposed methodology is the concept of social welfare because the interests of the consumers, generators and transmission operators have been accounted for simultaneously. The overall methodology has been implemented on a modified IEEE 5-bus test system (see Fig. 1) with the following assumptions:

1) The demand at all buses is defined by the load curve *LDC*. To incorporate seasonal variation of demand, the monthly *LDC*, in terms of the average monthly demand, has been adopted. Thus there are total of 12 possible load scenarios [19].

2) New generators are present at buses 6 and 7 and generator at bus 1 is taken as slack

generator [12].

3) FOR of all generators are specified [12]. At least two of generators should be online to inject power into the network[4].

4) All the probable new transmission lines are known *a priori* [12], while the capacity of the new transmission lines would be calculated by the planning procedure.

5) Length, impedance and FOR of transmission lines are specified [12].

6) The investment costs of lines, generators and the *EDNS* costs are taken from [12], [13].

### 2.1. Objective Function

The objective function **J** includes a sum of the total expected cost resulting from *DNS*, *GNS*, *WL* and investment for setting up the new transmission lines and generation capacities:

$$J = EC + T_{inv} + G_{inv} \tag{1}$$

$$EC = EDNS_{cost} + EGNS_{cost} + EWL_{cost} \tag{2}$$

Where, $T_{inv}$ and $G_{inv}$ are respectively the investment in setting up transmission lines and generators. The constraints related to the objective function are:

$$\sum_{g=1}^{NG} G(t)_g + \sum_{d=1}^{ND} D(t)_d = 0 \tag{3}$$

$$P_s = \sum_{j=1}^{b} \frac{1}{x_{sj}} (\delta_j - \delta_s) \tag{4}$$

Equation (3) ensures a balance between supply and demand, while equation (4) represents the DC-load flow formulation whose solution provides the power flow in all transmission lines. The definition of the terms in equations (1) and (2) are given below:

$$EDNS_{cost} = 730 \sum_{t=1}^{12} C_{EDNS}(t) * EDNS(t) \tag{5}$$

---

[4] In general, the minimum number of online generators is a function of the network size.

$$EGNS_{\text{cost}} = 730\sum_{t=1}^{12}\left(\begin{array}{c}C_{EGNS}(t)*EGNS(t)\\ +\sum_{s=1}^{g}C_{rl(s)}*EGO(t)_s\end{array}\right) \quad (6)$$

$$EWL_{\text{cost}} = 730\sum_{t=1}^{12}C_{EWL}(t)*EWL(t) \quad (7)$$

In equations (5) to (7), 730 refers to the number of hours in a month. *EGO(t)s* indicates expected outage when a generator at bus s is cut-off from the network during the simulations. In contrast, EGNS refers to unutilized generation capacity from generators connected to the network. *CEDNS, CEGNS, CEWL*, Crl(s) are the costs of *EDNS, EGNS, EWL*, and revenue loss of sth generator due to outage respectively, in units of *k$/MWh* respectively. The transmission and generation investment costs can be calculated from the following equations:

$$T_{inv} = \sum_{q=1}^{N_n} C_{T1(q)}*TL_q + \sum_{p=1}^{N_T} C_{T2(p)}*TL_p*F_p*OCF_p \quad (8)$$

$$G_{inv} = \sum_{k\in N_G} C_{G1(N_G)}*GC_{N_G} + 730\sum_{k=1}^{NG} C_{G2(NG)}*\sum_{t=1}^{12} PG_{NG} \quad (9)$$

$C_{T1(q)}$ is the capital investment cost of $q^{th}$ transmission line [13], achieved by curve fitting

$$C_{T1(q)} = 0.35 * F_q + 0.19 \quad (10)$$

and $TL_q$ is the length of $q^{th}$ transmission line (*km*), $N_T$ denotes total number of transmission lines in the network, $N_n$ represents number of new transmission lines, $F_q$ belongs to the capacity of $q^{th}$ transmission line (MW), $F_p$ is the capacity of $p^{th}$ transmission line (MW), and $C_{T2}$ is the annual operating and maintenance cost of line (k$/MW/km). $OCF_p = \dfrac{(1-FOR_p)}{FOR_p}$ is the operating cost factor of $q^{th}$ transmission line computed by forced outage rate (FOR), based on climatic and geographical conditions. $TL_q$ designates the length of $q^{th}$ transmission line (km), NG indicates number of all generators in the power system, $N_G$ refers to number of new generators (bus-6 and bus-7), [5] $C_{G1(N_G)}$ symbolize the capital cost of $N_G^{th}$ generator (k$/kW), $C_{G2(NG)}$ stand for operating and maintenance cost including fuel cost of NG$^{th}$ generator, (k$/kWh) [20], including fuel cost, $GC_{N_G}$ defines the proposed generation capacity before TEP at $N_G$ bus, and $PG_{NG}$ characterize the generated power at bus NG optimally for given load scenario.

---
[5] This cost is constant because generation capacity is defined at all generator buses before TEP.

### 2.2. Calculation of DNS, GNS, and WL

First a DC-load flow is run on the electrical network with the specified demand at each node (D$_S$) assuming that all transmission lines do not have capacity constraints. DNS at each node (DNS$_s$) is then calculated using the procedure given below.

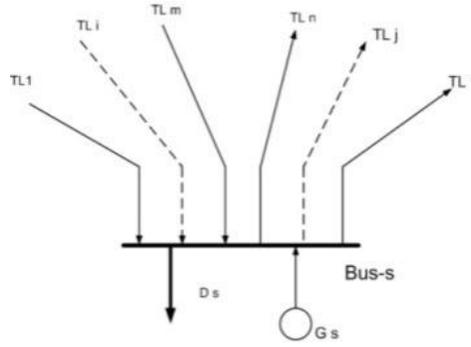

Fig. 2. Incoming and outgoing transmission lines at bus 's'

Assume there are "m" lines carrying power in to node "s" and "n" lines taking power out away from node "s". $DNS_s$ and $GNS_s$ are defined as (see Fig. 2):

$$DIFF_s = D_s - \min\left(\sum_{i=1}^{m}\left(T_{f,i,s}, T_{c,i,s}\right)\right) + \min\left(\sum_{j=1}^{n}\left(T_{f,j,s}, T_{c,j,s}\right)\right) - G_s \quad (11)$$

$$DNS_s = DIFF_s : if\ DIFF_s > 0 \quad (12)$$

$$GNS_s = abs(DIFF_s) : if\ DIFF_s < 0 \quad (13)$$

Where, $D_s$ and $G_s$ respectively stand for demand and generation at bus s, $T_{f,i,s}$ and $T_{f,j,s}$ are the flow on $i^{th}$ incoming and $j^{th}$ outgoing transmission line respectively connected to bus s, and $T_{c,i,s}$ and $T_{c,j,s}$ respectively the transmission capacity of $i^{th}$ incoming and $j^{th}$ outgoing transmission line

connected to bus *s*. Where the system *DNS*[6] and *GNS* are calculated as

$$DNS = \sum_{s=1}^{b} DNS_S \text{ and } GNS = \sum_{s=1}^{b} GNS_S \qquad (14)$$

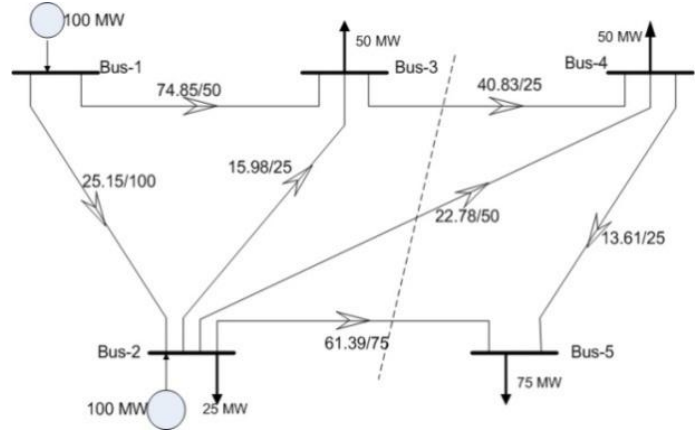

Fig. 3. Power flow in the network to illustrate the methodology for calculating $DNS_s$ and $GNS_s$.

Clearly $GNS_s$ is applicable to either a pure generator or a mixed (generator + demand) bus. The above concept can be understood by the example shown in Fig. 3, where, for example, 25.15/100 refers to power flow of 25.15 MW between buses 1 and 2, through a 100 MW capacity transmission line. Fig. 3 shows that two transmission lines connected from bus-1 to bus-3 and from bus-3 to bus-4 are overloaded. In fact, the overloading affects the power injection at generation bus as well as demand served at load buses. $DNS_s$ and $GNS_s$ computed by equation (11), (12) and (13) are shown in Table-II.

TABLE II
CALCULATION OF $DNS_S$ AND $GNS_S$ IN THE NETWORK SHOWN IN FIG. 3

| Bus | *DIFF*(MW) |  |
|---|---|---|
| 1 | -25 | $(DIFF_1<0) \Rightarrow GNS_1 = 25$ |
| 2 | 0 | $(DIFF_1=0) \Rightarrow GNS=DNS=0$ |
| 3 | 9.02 | $(DIFF_3>0) \Rightarrow DNS_3 = 9.02$ |
| 4 | 15.83 | $(DIFF_4>0) \Rightarrow DNS_4 = 15.84$ |
| 5 | 0 | $(DIFF_1=0) \Rightarrow GNS=DNS=0$ |
|  | $DNS = 24.86\text{MW} \cong GNS$ | |

---

[6] Subscript 's' denoted that the quantity is calculated at bus, while the absence of subscript 's' tends for aggregated calculation thus called system DNS and system GNS.

From the above table it can be observed that *GNS* is equal to the *DNS* in the network. The computed $GNS_s$ and $DNS_s$ are only associated with those buses which are connected to the power system through constrained transmission lines. The $GNS_s$ at bus can be defined as the loss to generators, which has to shutdown due to constrained network. Wheeling loss is computed by considering power flow in congested lines:

$$WL = \sum_{k \in Congested} (T_{f,k} - T_{c,k}) \tag{15}$$

Where, $T_{f,k}$ is the power flow and $T_{c,k}$ is the capacity of the $k^{th}$ congested transmission line.

The aggregated wheeling loss is the sum of wheeling loss on all k congested transmission lines. As shown in Fig. 3, transmission lines between buses 1 and 3 and buses 3 and 4 are congested and demand more capacity to wheel power. Thus, the *WL* in the example system given in Fig. 3, is (27.85+15.83=43.68 MW). Multiplication of *WL* with cost of wheeling per MW is cost of wheeling loss belongs to the transmission system owner. The computations of *DNS*, *GNS* and *WL* are repeated multiple times for different demand scenarios and contingencies generated by *MCS*. Thus, from the distributions of *DNS*, *GNS* and *WL* the associated *EDNS*, *EGNS* and *EWL* are calculated. Equation (16) defines *EDNS*:

$$EDNS = \sum_{i \in dDNS}^{b} DNS_i * pd_i \tag{16}$$

Here, $DNS_i$ represents $i^{th}$ sample of the distribution of *DNS* and $pd_i$ belongs to probability of occurrence of $DNS_i$. The definitions of *EGNS* and *EWL* are similar to that of *EDNS* as shown in equation (16), with $DNS_i$ being replaced by $GNS_i$ and $WL_i$ and their corresponding probabilities respectively.

### 2.3. *Monte Carlo simulation*

In Monte Carlo simulation (*MCS*), random variables are drawn separately for each of the transmission lines under consideration. All lines with variable values between FOR and 1 are retained for power transmission. If the value of the variable is between 0 and FOR then the line is assigned "zero" capacity, that is, it is omitted from the network. Subsequently, the reduced network is used for calculating *DNS* and *GNS*.

During *MCS*, situations may arise which can lead to (i) the isolation of a demand bus or/and (ii) the specified power not being evacuated from a generator, resulting in "generation not served." An example of such a case is show in Fig. 3, where the generation at buses 1 and 2 match the sum of the loads at buses 3, 4, and 5. If the transmission lines connected to bus 2-4, bus 2-5, and bus 3-4 are removed during *MCS*, the demand buses 4 and 5 are isolated from the network. This also results in a situation where the total generation does not match the total demand in the network, which in turn, is not amenable to power flow analysis for calculating *DNS* and *GNS*. To ensure that these situations are eliminated from consideration, the following constraints are imposed:

$$0 < DNS(t) < D(t) \quad \text{or} \quad DNS(t)_s < D(t)_s \qquad (17)$$

$$0 < GNS(t) < G(t) \quad \text{or} \quad GNS(t)_s < G(t)_s \qquad (18)$$

In equation (17) and (18), *D* and *G* refer to total demand and total generation in the network respectively, where $t^7$ represents the time instant for which simulation is carried out.

### 2.4. Roulette Wheel

After each DC-load flow run, all congested lines, where power flow is greater than their respective transmission capacities, are assigned the value 1. Non-congested lines are assigned a value of 0. The probability of congestion of each transmission line in the network is computed, while after each DC-load flow run after contingency

---

[7] This is associated with the time instant of load curve, in our simulation 12 time points are taken to encounter 12 load scenarios.

$$P_{con,j} = \frac{N_{1,j}}{N_{MCS}} \tag{19}$$

*Where, $P_{con,j}$ is the probability of congestion and $N_{1,j}$ is the number of "1" for the $j^{th}$ line.*

Now, an imaginary roulette wheel [21], [22] is constructed where each transmission line in the network is represented by a separate segment. The area of the segment corresponding to each transmission line is proportional to its probability of congestion. The wheel is rotated N times, where N is the number of transmission lines having a congestion probability greater than 0.1– both existing and proposed -- in the network. At the end of each rotation, the segment at which the pointer stops, and the corresponding transmission line, is noted. The capacity of each transmission line at the end of N rotations is updated according to

$$F_j = F_j^o + m_j \Delta F_j \tag{20}$$

*Where, $F_j^0$ is the base capacity of the $j^{th}$ transmission line, $m_j$ is the number of times out of N that the roulette pointer stops at the segment corresponding to the $j^{th}$ transmission line and $\Delta F_j$ is the value by which the transmission capacity is increased in the $j^{th}$ transmission line.*

It is important to ensure that the transmission capacities are not over-specified because it would result in superfluous investment. That is, the transmission capacities should be such that it can withstand a reasonable level of overloading without affecting system stability and security. In contingency states, overloading of transmission lines for short-term emergency rating can be tolerated by a factor of 1.1 to 2 from the nominal rating, and for long-term emergency rating the value of factor can be from 1.05 to 1.8 [23]. These levels of overloading can be handled for 15 to 30 minutes by the safety instruments and load shedding strategies. In our calculations, we have specified a maximum overloading of 10%. Consequently, the process of capacity updating of a transmission line –elaborated in a later section – is terminated, once the probability of congestion goes below 0.1. After the roulette wheel is rotated $N_T$ times, the marginal expected cost (*MEC*) and marginal investment (*MI*) are calculated from equation (21) and (22) respectively:

$$MEC = \frac{d(EC)}{dF_N} = \frac{EC^i - EC^{i-1}}{F_{N,i} - F_{N,i-1}} \qquad (21)$$

*Where, $EC^i$* is expected cost in $i^{th}$ iteration, $EC^{i-1}$ is expected cost in $(i-1)^{th}$ iteration, $F_N$ is updated network capacity of the network in one complete simulation of roulette wheel selection and $dF_N$ is change in the capacity of the network

$$MEC = \frac{d(T_{inv})}{dF_N} = \frac{T_{inv}^i - T_{inv}^{i-1}}{F_{N,i} - F_{N,i-1}} \qquad (22)$$

Where, $T_{inv}^i$ is transmission investment in $i^{th}$ iteration[8] and $T_{inv}^{i-1}$ is transmission investment in $(i-1)^{th}$ iteration. When *MEC* equals *MI*, an optimal solution has been found where the total cost is minimized.

### 3. ALGORITHM

The proposed probabilistic TEP algorithm is described below where Genetic algorithm (GA) is used as an optimization model [24], [25]:

1) Generate initial population of chromosomes having the length equal to the set of proposed lines, where 1 and 0 represent the active and inactive lines respectively. All existing lines are automatically considered as active.

2) Take one chromosome from the population for simulation and assign a minimum transmission capacity of 5 MW ($F_i^0$) to each active proposed transmission line. Compute associated total investment.

---

[8] In the proposed methodology TEP is carried out after planning of generators thus generation investment in constant.

3) Take scenario 1 out of 12 scenarios generated from *LDC*.

4) Generate a string by removing some transmission lines from the chromosome, as well as generate the string of off-line generators using *MCS*.

5) Check islanding condition. If system is in islanding mode, go to Step 4.

6) If slack generator is removed during *MCS*, replace the slack generator with the next higher capacity generator. Dispatch the generators according to merit-order dispatch to match the total demand.

7) Run DC-load flow with remaining lines. Store the power flow in all active lines.

8) Go to step 4, and repeat for multiple contingencies (transmission and generation) in *MCS*.

9) Go to step 3, repeat for all scenarios generated from *LDC*.

10) Calculate *EDNS, EGNS* and *EWL* and the corresponding *EC*.

11) Compute the marginal expected cost (*MEC*) and marginal investment (*MI*).

12) If *MEC* is equal to *MI*, then go to step 18.

13) Compute the probability of congestion for each active line, using data from stored power flow data. Active lines with $p_{con,j} \leq 0.1$ are not updated.

14) Create roulette wheel with each active line having $p_{con,j} \geq 0.1$.

15) Update transmission capacities according to procedure specified in section *2.4*. $\Delta F_j$ is taken as 5 MW[9].

16) Calculate the total investment cost in the updated network.

17) Go to step 10.

18) Go to step 2.

19) Apply GA and find the least cost solution using the objective function in equation (1).

---

[9] Smaller values of $\Delta F_j$ should give a more precise solution. Calculations show that changing $\Delta F_j$ from 5 MW to 1 MW affects *EDNS* and *EGNS* by only 10%.

The corresponding flow chart is shown in Fig. 4.

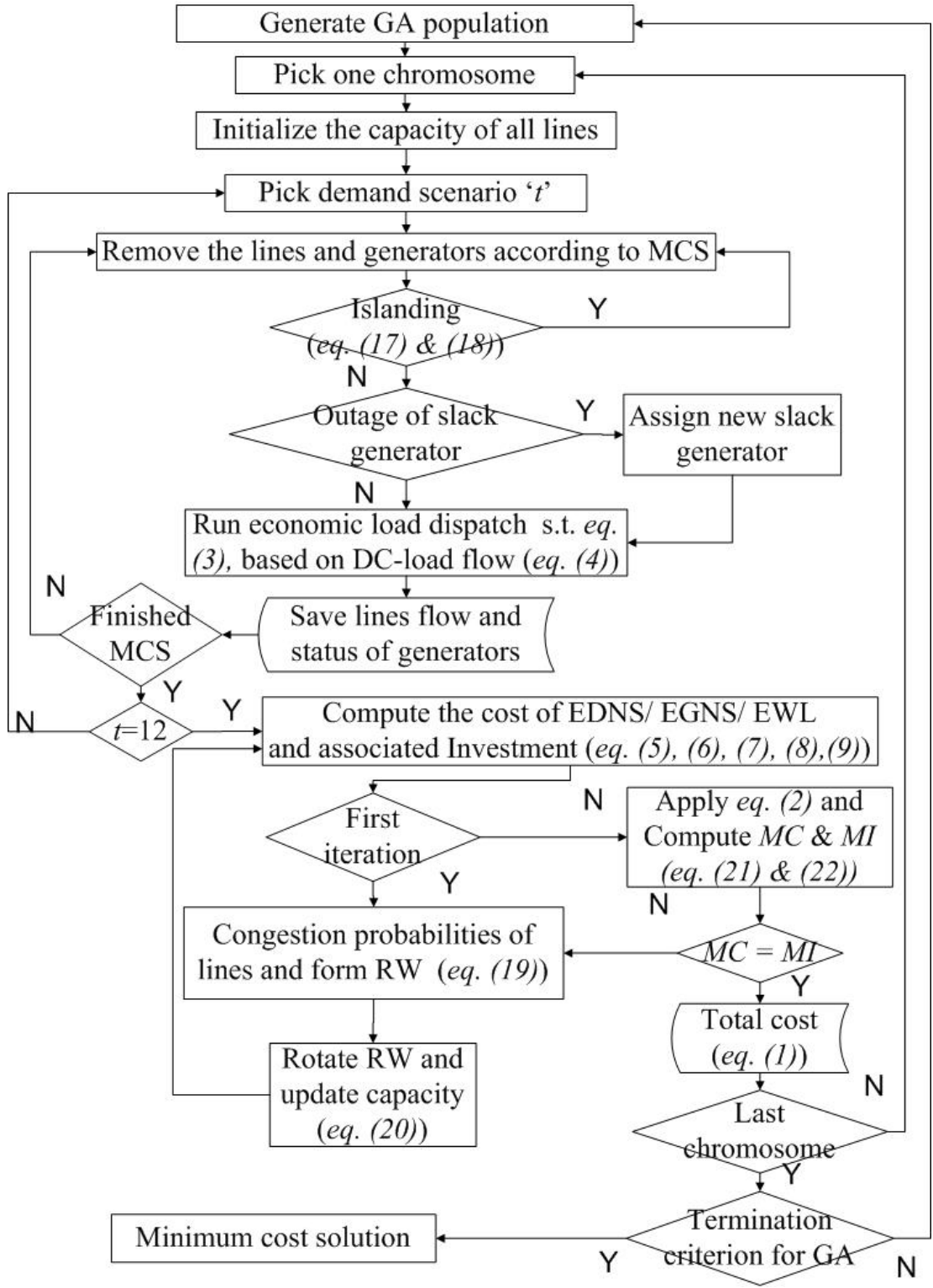

Fig. 4. Flow chart of the proposed TEP methodology.

## 4. CASE STUDY

The implementation of the intrinsic methodology has been carried out on the 5-Bus IEEE test system shown in Fig. 1. The proposed algorithm has been used to identify a set of optimal transmission lines and their respective capacities from the probable new transmission lines [12]. In Fig. 1, BD represents bus demand and TL represents the transmission line. The numbers in front of BD and TL represent the corresponding bus and transmission line numbers. Bus-6 and bus-7 are the new generator buses, while the bus-1 is the slack bus. Table II depicts both the existing (TL 1 to TL 7) and the possible new (TL 8 to TL 21) transmission lines. To find the optimal solution 450 generations for GA are carried out. In each generation 30 chromosomes have been taken. To encounter most of the contingencies in the planning process, 1000 iterations of *MCS* is carried out [10]. All existing transmission lines and alternative new lines with their initial capacities have given in Table II.

There are two possible options for TEP: (i) add only new transmission lines to the existing network and (ii) Upgrading capacity of existing lines in addition to putting up new lines. To determine which of the above two option yields the minimum cost, *J*, the TEP methodology outlined in section *2* was carried out with a single demand scenario [12]. Table II also shows the optimal set of new transmission lines and their associated capacities in *NL* column. Please note that here the existing transmission lines have not been updated. From Table II we can observe that the nine transmission lines are selected optimally with an *EDNS/ EGNS* of 71.43 MW, which is 17.85% of the total demand. The *EWL is* 84.26 MW. Calculations showed that further increase in the capacities of the new lines shown in Table II, column *NL* did not significantly change *EDNS* and *EWL*.

---

[10] From the simulations we have observed that for the given network, 1000 iterations of *MCS* result are sufficient for convergence. In reality, the number of iterations depends on the number of transmission lines (existing as well as new) in the power system.

TABLE II

DETAILS OF TRANSMISSION LINES IN THE NETWORK, SHOWN IN FIG. 1, USED IN THE CASE STUDY
(RESULT FOR SINGLE DEMAND SCENARIO AS GIVEN IN [12])

|  | Length (km) | *frombus-tobus* | Existing Capacity | Updated Capacity (MW) NL[11] | WEL[12] |
|---|---|---|---|---|---|
| TL 1 | 40 | 1-2 | 100 | 100 | 100 |
| TL 2 | 30 | 1-3 | 50 | 50 | 70 |
| TL 3 | 25 | 2-3 | 25 | 25 | 85 |
| TL 4 | 45 | 2-4 | 50 | 50 | 60 |
| TL 5 | 65 | 2-5 | 75 | 75 | 125 |
| TL 6 | 20 | 3-4 | 25 | 25 | 95 |
| TL 7 | 20 | 4-5 | 25 | 25 | 75 |
| TL 8 | 50 | 1-4 |  | 145 | 135 |
| TL 9 | 70 | 1-5 |  | - | 55 |
| TL 10 | 15 | 1-6 |  | 60 | - |
| TL 11 | 80 | 1-7 |  | 95 | 90 |
| TL 12 | 50 | 2-6 |  | 65 | 65 |
| TL 13 | 40 | 2-7 | [13]Cap | 130 | - |
| TL 14 | 40 | 3-5 |  | - | 110 |
| TL 15 | 35 | 3-6 |  | 95 | 85 |
| TL 16 | 50 | 3-7 |  | - | - |
| TL 17 | 55 | 4-6 |  | 65 | 60 |
| TL 18 | 30 | 4-7 |  | 140 | 135 |
| TL 19 | 70 | 5-6 |  | 155 | - |
| TL 20 | 15 | 5-7 |  | - | 140 |
| TL 21 | 60 | 6-7 |  | - |  |
|  |  |  | *EDNS* (MW) | 71.43 | 20.71 |
|  |  |  | *EWL* (MW) | 84.26 | 26.13 |
|  |  |  | $T_{inv}$ *(M$)* | 19.3 | 22.4 |
|  |  |  | *EC (M$)* | 55.4 | 16.59 |
|  |  |  | [14]*J (M$)* | 140.4 | 104.69 |

This suggests that further reduction in *EDNS, EGNS* and *EWL* is possible only by removing congestion in the existing lines. For example, consider bus-2 in Fig. 1. Here we can see that congestion in lines TL-3 and TL-4 cannot be removed by increasing capacity of TL-13. Consequently, simulations have also been carried out for the scenario where all lines in the network, both the existing and the new probable lines, were considered for capacity up-gradation. Column *WEL* in Table II shows results when all lines, both existing and new lines, are considered for capacity up-gradation. A comparison of column *NL* and *WEL* in Table-II shows that *EDNS* has been reduced by 71% to 20.7 MW, which, in turn, is 5.2% of the total demand. Similarly, there is a 69% reduction in *EWL*. The reduction in *EDNS* and *EWL* has primarily been due to the higher capacities of six out of the seven existing lines. At this point we can conclude that the achieved result is economic with an acceptable reliability level, where a 16.06% increase in the network investment ($T_{inv}$) has reduced the *EC* by 70.05% (thus objective function, *J* by 25.4%). Detailed simulations with the LDC have therefore been carried out by updating old lines along with new lines. Updating of old lines may be either through re-conductering or adding parallel lines. Parallel transmission lines improve the reliability of the electrical network; parallel lines are proposed where the capacity approximately doubles.

---

[11] Updated capacity of the network without updating of old lines.
[12] Updated capacity of the network including updating of old lines.
[13] Initial capacity of all lines is taken 5 MW.
[14] Since generator capital investment is constant thus does not show in *J*.

For relatively lower capacity update, upgrading of the existing line conductor is preferable due to economics of the power systems. Table-III compares transmission expansion between the deterministic *N-2/N-1* and the probabilistic *MCS* contingency methodologies. Here it is noted that *N-1* and *N-2* planning is carried out for peak demand scenarios while *MCS* follows average load curve. As expected, a comparison of columns of Tables III shows that for *N-1* and *N-2* based TEP, *EDNS, EGNS, EWL* and the overall costs are lower than TEP based on *MCS*. Of course, the extent of the difference will depend on the magnitude of EDNS, EGNS, and EWL costs vis-à-vis transmission investment costs. Because of the issue of cost versus reliability, an important question arises: Should TEP be based on *MCS* or *N-1/N-2* contingency analyses? For developing countries such as India, where the realization of its economic growth potential is critically dependent on a reliable electricity supply network, TEP based on *MCS* may be the preferred methodology. The additional costs incurred in implementing this methodology will be compensated several times over by the expected increase in the GDP. A rigorous analysis of this question, which will include congestion costs, will be the subject of a forthcoming paper

TABLE III
DETAILS OF TRANSMISSION LINES IN THE NETWORK FOR DIFFERENT CONTINGENCIES CASES AND DEMAND SCENARIOS

|   | MCS | N-1 | N-2 |   |
|---|---|---|---|---|
| TL 1 | 100 | 100 | 100 | Updated Capacity |
| TL 2 | 70 | 50 | 60 | |
| TL 3 | 50 | 35 | 35 | |
| TL 4 | 50 | 50 | 50 | |
| TL 5 | 75 | 75 | 75 | |
| TL 6 | 95 | 75 | 100 | |
| TL 7 | 50 | 35 | 40 | |
| TL 8 | 95 | 100 | 105 | |
| TL 9 | 75 | - | - | |
| TL 14 | 95 | 85 | 95 | |
| TL 15 | 90 | - | - | |
| TL 16 | 115 | 65 | 110 | |
| TL 17 | 90 | 85 | 90 | |
| TL 18 | 120 | 115 | 130 | |
| TL 19 | 100 | 90 | 95 | |
| TL 21 | 100 | - | 40 | |
| EDNS | 19.09 | 11.39 | 14.44 | |
| EWL | 21.56 | 13.72 | 16.97 | |
| $T_{inv}$ | 19.18 | 11.71 | 14.76 | |
| EC | 14.8 | 8.9 | 11.3 | M$ |
| J | 98.23 | 84.91 | 90.36 | |

## 5. CONCLUSION

A new methodology for transmission expansion planning that (i) does not require *a priori* specification of new/additional transmission capacities, (ii) uses the concept of social welfare, (iii) uses load flow and electrical laws to calculate EDNS, and (iv) incorporates the effect of seasonal variation through the LDC has been presented. Two new concepts have been introduced in this paper. First, load flow has been used to calculate DNS, compared to the conventional min-flow max-cut approach that does not necessarily follow electrical laws. Second, the roulette wheel model has been used to calculate the new/additional transmission capacities. In fact, the roulette wheel methodology is applicable to both deterministic and probabilistic contingency analyses based TEP.

Simulations show an important result: addition of only new transmission lines is not sufficient to minimize the magnitude of unfulfilled demand (*EDNS*). In fact, calculations show that a further 71% reduction in *EDNS* was possible primarily due to the capacity up-gradation of six out of the seven existing lines. Results also show that for *N-1* and *N-2* based TEP, which uses peak LDC, the overall costs are lower than the *MCS* based TEP, which uses the average LDC.

To make the proposed methodology more realistic, several important aspects of power system need to be included, for example, reactive power costs and costs related to carbon credits. Clearly, application of the proposed methodology for large networks can be cumbersome, which will be dealt with in a forthcoming paper. To reduce the computation time, a methodology would be proposed to (i) reduce the number of probable new transmission lines and (ii) replace multiple load flow by the generalized line outage distribution factor (GLODF).